# Orange Quality Grading with Deep Learning

Mohamed Lamine Mekhalfi [1], Paul Chippendale [1], Francisco Fraile [2], Marcos Rico [3]

[1] *Fondazione Bruno Kessler, Via Sommarive, 18, 38123, Trento-Italy.*
[2] *Universitat Politècnica de Valencia, Camino de Vera, s/n 46022, Valencia-Spain.*
[3] *Multiscan Technologies, S.L, C. la Safor, 2, 03820 Cocentaina, Alicante, Spain.*

**Abstract.** Orange grading is a crucial step in the fruit industry, as it helps to sort oranges according to different criteria such as size, quality, ripeness, and health condition, ensuring safety for human consumption and better price allocation and client satisfaction. Automated grading enables faster processing, precision, and reduced human labor. In this paper, we implement a deep learning-based solution for orange grading via machine vision. Unlike typical grading systems that analyze fruits from a single view, we capture multiview images of each single orange in order to enable a richer representation. Afterwards, we compose the acquired images into one collage. This enables the analysis of the whole orange skin. We train a convolutional neural network (CNN) on the composed images to grade the oranges into three classes, namely 'good', 'bad', and 'undefined'. We also evaluate the performance with two different CNNs (ResNet-18 and SqueezeNet). We show experimentally that multi-view grading is superior to single view grading.

## 1. Introduction

Fruit grading is an essential process that ensures the quality of the produce and enhances its marketing efficiency in the food industry. Grading refers to the sorting or categorization of fruit items into different grades according to several attributes such as size, shape, color, weight and volume. It helps to remove undesirable or foreign matters from the harvested crops into various fractions, which improves product uniformity within a particular grade and serves as the basis for determining price and suitable customers.

In this regard, automated grading offers several advantages with respect to manual human grading, upsides include **(i) Speed and productivity**: Automated grading solutions typically grade the food items much faster than humans, increasing efficiency and keeping labor costs in check. Further, this is subject to boost productivity and economic growth in the fruit industry, **(ii) Accuracy**: Recent grading technologies can measure and grade fruits and vegetables at a comparable or even higher accuracy than human operators, **(iii) Consistency**: Owing to their high accuracy, automated grading solutions can maintain consistency and uniformity in grading, which is difficult to achieve with manual inspection, **(iv) Reduced waste**: Automated grading solutions can help reduce waste by identifying and removing damaged or defective produce before it is packaged, **(v) Increased safety**: In some scenarios, human graders may have to operate within a cold, tight, noisy and crowded industrial setup with heavy loads of machinery and produce, which likely puts their safety at stake. Automated grading surpasses this bottleneck, **(vi) Persistence**: Grading is a repetitive task, yet the quality of human grading decays over the working hours due to factors like tiredness, lack of attention and perhaps boredom. Automated grading is not prone to these downsides, **(vii) Customization**: Automated grading solutions can often be upgraded according to different grading criteria or tailored to a different product (e.g., switching from oranges to tomatoes), **(viii) Data history**: They offer the property of collecting data streams of fruits and vegetables passing through the grading system. These data represent a valuable asset that can be exploited to upgrade the grading system in terms of either hardware or software for a superior performance.

Emails: mmekhalfi@fbk.eu; chippendale@fbk.eu ; ffraile@cigip.upv.es; mrico@multiscan.eu.

On the other hand, two disadvantages that may characterize automated fruit grading include (i) initial cost outlay, (ii) more energy consumption and (iii) regular maintenance to prevent eventual breakdowns. However, the impact of these downsides is not tangible comparable to the advantages listed above. Further backup plans are ought to compensate for any eventual underperformance.

## 2. Related work

The literature concerned with fruit grading has developed interesting contributions so far [1]. Here, we emphasize mainly on vision-based solutions as they constitute the scope of this paper, besides the fact that they represent the bulk of the recent state-of-the-art. For instance, in [2] a vision system was developed to grade tomatoes into four classes, namely defective or non-defective, and ripe or unripe. The components of the system consist of a conveyor belt that transports the tomatoes, an optical camera to acquire images of the passing fruits and a processing unit that incorporates an Artificial Neural Network (ANN) for the classification of the fruits. Spectrophotometry and machine learning were explored for Apple fruit grading in [3], where a 72% accuracy was achieved on unseen fruits. In [4], traditional image processing techniques are combined with an ANN for banana grading with an overall accuracy of 97%. In [5], apple grading was addressed via simple pixel-level thresholding techniques. However, thresholding-based solutions are highly prone to performance drops if lighting conditions are altered. For more insights on relevant literature, the reader is referred to [6-7].

It can be noted that a big deal of previous works which are concerned with food grading and classification remain dependent on typical handcrafted image processing pipelines, which poses generalization challenges to new domains (e.g., if the acquisition sensors are replaced with different ones, or if the color of the background is changed). Therefore, recent contributions tend to apply deep learning techniques owing to their cutting-edge performance and real-time speed (thanks to powerful graphical processing units). For instance, in [8] 3D meshes of apple fruits are acquired by means of a 3D sensor, and a Convolutional Neural Network (CNN) model is developed to grade apples into bruised and healthy categories at an accuracy of 97.67%. RGB and hyperspectral imaging were combined to grade banana fruits into three classes by means of a CNN in [9]. In [10], an optical sensor was adopted to capture images of Green Plums that are exposed to LED lighting. Afterwards, a CNN model was exploited to grade them into five classes, namely, rot, spot, scar, crack, and normal. The work in [11] compares two deep learning architectures (i.e., AlexNet and VGG16) for classifying Jujube fruits into three classes according to their maturity level (i.e., unripe, ripe, and over-ripe). VGG16 yielded an accuracy of 98.41% whilst AlexNet scored 96.21%.

Regarding the vision aspect of the aforementioned solutions, one can observe that they mostly capture only one side (i.e., one image) of the subject fruit being graded, overlooking the fact that the unseen sides of the fruit may represent precious information that can help improve the grading performance. To tackle this, in this work we propose a multi-view system for grading orange fruits based on RGB imaging. In particular, each single fruit is rolled several times, while RGB images are acquired each time. Next, the acquired image instances are put together to form a grid of images that pertain to the same fruit. We develop a deep learning grading pipeline to grade each orange fruit into one of three classes, namely 'good', 'bad', and 'undefined' and assess the performance of our solution with respect to each class.

## 3. Dataset

The dataset was acquired by Multiscan Technologies, S.L (Pol. Ind. Els Algars C/ La Safor, 2 03820 Cocentaina, Alicante – Spain). The dataset contains the images along with their grade annotations. To this end, the oranges go through a roller conveyor that moves them forward and rotates them simultaneously (See Fig. 1). As depicted in Fig. 2, each orange fruit is captured from different viewpoints.

The oranges were captured by means of Sony IMX429 camera, which is placed in a top-down view at approximately 1 meter from the fruit plane. The oranges are exposed to Cool white LED as a uniform lighting source.

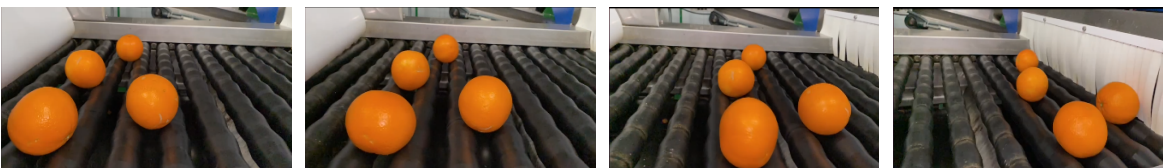

**Fig. 1.** Several instances of oranges on a roller conveyor.

Fig. 2 also shows instances from the three grading classes. The 'good' grade oranges manifest a clean skin, while the 'bad' class oranges often contain blemishes and bruises of various severities on the outside. The 'undefined' class oranges, however, report imperfections that are neither too grave to be categorized as 'bad' nor insignificant to be considered 'good'. Note that the notations 'good, bad, undefined' as well as 'bueno, malo, indefinido' are used interchangeably in this paper.

The dataset totals 452 orange samples (each sample features several views of the same orange) with highly varying sample distribution per class. We divide the dataset into two subsets, namely a training set of 317 samples (70% of the dataset) and 135 test samples (yet 30% of the total count). The statistics are given in Table 1.

**Table 1.** Statistics of the orange dataset.

|  | Training | Test |
| --- | --- | --- |
| Bueno | 78 | 33 |
| Malo | 206 | 88 |
| Indefinido | 33 | 14 |
| Total | 317 | 135 |

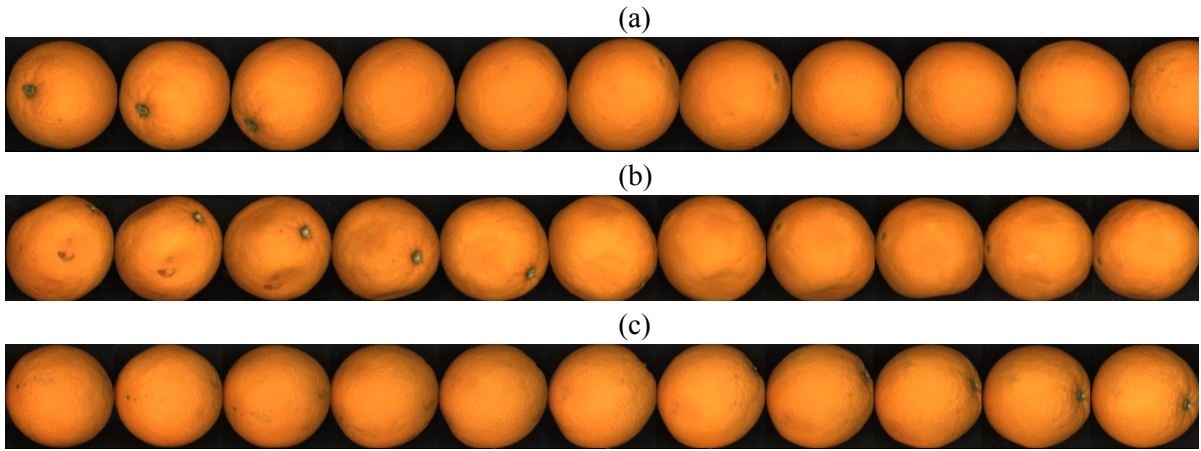

**Fig. 2.** Samples from each class. (a) Good (Bueno), (b) Bad (Malo), and (c) Undefined (Indefinido).

## 4. Experiments

As mentioned earlier, the main factor characterizing each of the three classes is the presence as well as the size of bruises and blemishes on the skin of the oranges. Therefore, to ensure a better visibility of these latter, multiple views of each orange are combined together to form one single RGB image as shown in Fig. 2.

Regarding the classification task, we adopt two deep learning models. The first one is based on ResNet-18 architecture, which has proven useful for image classification [12]. However, the grading solution may eventually be deployed in real-time, which may require a deep model with fewer parameters. Thus, we also opt for a second classification model based on SqueezeNet [13]. The images are resized to dimension 2500x300 before being fed to the aforementioned models.

The performance measure is expressed in terms of classification Accuracy per class, which indicates the sum of correctly classified samples over the total number of available samples. We also report the average classification Accuracy across the three classes, and the Overall classification accuracy which represents a classification Accuracy regarding the samples of all the classes put together.

First, we report the scores when all the orange views are considered in Table 2. It can be observed that the class 'Malo' reports the highest score, followed by 'Bad' class and then 'Indefinido' class, which is due to the imbalance in the number of samples per class as summarized in Table 1. In particular, the class 'Indefinido' contains very few samples, and this reflects a low classification rate. On the other hand, the overall classification is plausible, owing mainly to the high number of samples in the 'Malo' class that reports a high accuracy.

We also note that ResNet-18 performs better than SqueezeNet when classifying the 'Bad' class, while the opposite is true for the 'Bueno' class. This is perhaps due to the fact that ResNet-18 is a larger model that performs well when abundant images are available. In terms of average classification accuracy, SqueezeNet performs better than ResNet-18 (+2.7%). Both models perform almost on par in terms of overall accuracy.

**Table 2.** Multiview classification scores (%).

|  | Bueno | Malo | Indefinido | Avg. | Overall |
| --- | --- | --- | --- | --- | --- |
| **ResNet-18** | 57.60 | 87.50 | 21.40 | 55.50 | 73.30 |
| **SqueezeNet** | 72.70 | 80.70 | 21.40 | 58.30 | 72.60 |

Second, in order to evidence the choice of Multiview orange classification, we also conduct a further experiment by retaining only one view of each orange while discarding the remaining ones. The scores are given in Table 3. Although the classification score of the 'Malo' class increases for both models as this class features a higher number of samples, the scores pertaining to the other two classes have dropped significantly. This highlights the advantage of multiview image analysis for orange grading.

**Table 3.** Single view classification scores (%).

|  | Bueno | Malo | Indefinido | Avg. | Overall |
| --- | --- | --- | --- | --- | --- |
| **ResNet-18** | 39.40 | 88.60 | 7.10 | 45.10 | 68.10 |
| **SqueezeNet** | 42.40 | 94.30 | 0.00 | 45.60 | 71.90 |

It is to note that the scores summarized in Tables 2 and 3 were obtained with models that were pretrained on the ImageNet dataset. This is a common practice in deep learning image analysis in order to transfer the knowledge learned by a model on a certain dataset (typically a large one like ImageNet) to another dataset (or task) like the one acquired within the scope of this paper. This is referred to as transfer learning and has been shown to incur improvements. Therefore, in a third experiment, we train both models from scratch (without pretraining on ImageNet) and report the results in Table 4. Interestingly, the scores of the class 'Malo' have improved drastically in view of the high number of samples in this class. Regarding the 'Bueno' class, ResNet-18 maintains its performance while SqueezeNet collapses, which may be due to the residual block mechanism in ResNet-18. The classification of the 'Indefinido' samples does not seem possible by both models. Therefore, for datasets with a few samples, pretraining is essential.

**Table 4.** Multiview classification scores (%) with scratch-trained models.

|  | Bueno | Malo | Indefinido | Avg. | Overall |
| --- | --- | --- | --- | --- | --- |
| **ResNet-18** | 57.60 | 90.90 | 0.00 | 49.50 | 73.30 |
| **SqueezeNet** | 0.00 | 100.00 | 0.00 | 33.30 | 65.20 |

As per the lower classification rates of the 'Bueno' and the 'Indefinido' classes with respect to the Malo' class, and apart from the dataset imbalance, it is to note that the annotation of the dataset plays a fundamental role. For instance, we depict two qualitative examples in Fig. 3., where the first row shows instances of an orange that was annotated as 'Bueno', and the second row illustrates instances of an orange that was annotated as 'Indefinido'. Visibly the 'Indefinido' orange looks smoother and mode uniform than the 'Bueno' one.

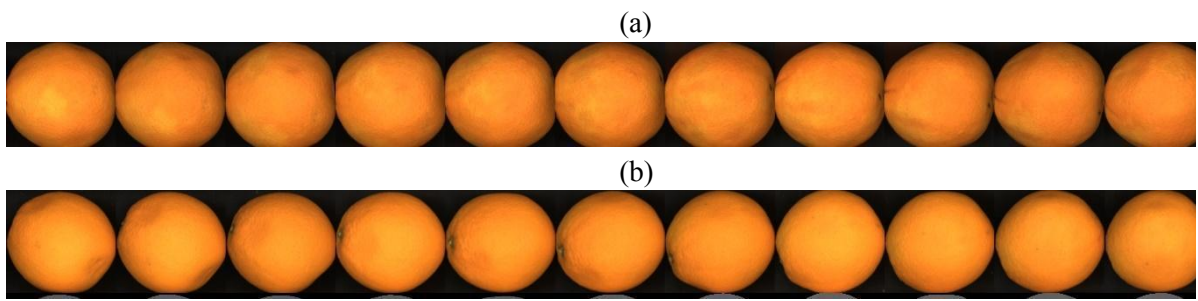

(a)

(b)

**Fig. 3.** Annotation examples. (a) Good (Bueno) sample, and (b) Undefined (Indefinido) sample.

## 5. Conclusion

In this paper, we addressed the problem of orange external quality grading into three classes by means of deep learning. Two classification models were applied, namely ResNet-18 and SqueezeNet. Overall, the classification scores are plausible except for the classes that feature few orange samples.

Future improvements include (i) increasing the size of the dataset, and (ii) in case the previous option is not affordable, alternative paradigms to deep learning can be implemented. For instance, traditional thresholding techniques can be developed in order to locate and estimate the size of blemishes in each orange view. This may require prior segmentation of oranges, which we believe it can be easily accomplished based on zero-shot models such as Segment Anything foundation model [14].

## 6. Acknowledgements

This paper was funded by the AGILEHAND project (Smart Grading, Handling and Packaging Solutions for Soft and Deformable Products in Agile and Reconfigurable Lines) under the European Union's Horizon Europe research and innovation programme under grant agreement No. 101092043.